\documentclass[a4paper,doc,natbib]{apa6}

\usepackage[english]{babel}
\usepackage[utf8x]{inputenc}
\usepackage{amsmath,amssymb}
\usepackage{bm}
\usepackage{graphicx}
\usepackage{subcaption}
\usepackage{caption}
\usepackage{url}
\usepackage[colorinlistoftodos]{todonotes}
\usepackage{float}
\usepackage{fancyhdr} 
\title{SoccerSynth Field: enhancing field detection with synthetic data from virtual soccer simulator}
\shorttitle{null}
\author{\textit{HaoBin Qin}\textsuperscript{\textit{*}}, \textit{ Jiale Fang}\textsuperscript{\textit{*}}, \textit{Keisuke Fujii}\textsuperscript{\textit{a},\textit{b}}}
\affiliation{
\textsuperscript{\textit{a}}\textit{Graduate School of Informatics, Nagoya University, Nagoya, Aichi, Japan;}
\\
\textsuperscript{\textit{b}}\textit{RIKEN Center for Advanced Intelligence Project, Tokyo, Tokyo, Japan;}
\\
\textsuperscript{\textit{*}}\textit{equal contributions}}

\abstract{
Field detection in team sports is an essential task in sports video analysis.
However, collecting large-scale and diverse real-world datasets for training detection models is often cost and time-consuming. Synthetic datasets, which allow controlled variability in lighting, textures, and camera angles, will be a promising alternative for addressing these problems.
This study addresses the challenges of high costs and difficulties in collecting real-world datasets by investigating the effectiveness of pretraining models using synthetic datasets.
In this paper, we propose the effectiveness of using a synthetic dataset (SoccerSynth-Field) for soccer field detection. A synthetic soccer field dataset was created to pretrain models, and the performance of these models was compared with models trained on real-world datasets. 
The results demonstrate that models pretrained on the synthetic dataset exhibit superior performance in detecting soccer fields. This highlights the effectiveness of synthetic data in enhancing model robustness and accuracy, offering a cost-effective and scalable solution for advancing detection tasks in sports field detection. }


\makeatletter
\renewcommand{\section}{\@startsection{section}{1}{\z@}%
  {-3.5ex \@plus -1ex \@minus -.2ex}%
  {2.3ex \@plus .2ex}%
  {\normalfont\Large\bfseries\raggedright}}
\makeatother
\begin{document}
\maketitle
\section{Introduction}
Field detection in team sports is an essential task in sports video analysis, enabling a more effective understanding of player movements, game dynamics, and strategy analysis. Traditionally, models for detecting fields have been trained in real-world datasets. However, obtaining large-scale and diverse real-world data can be expensive and time-consuming. For example, datasets used in broadcast sports analysis often require capturing a range of environmental conditions, different camera angles, and field variations, which is logistically challenging and costly. Many commonly used datasets, such as WorldCup 14 \citep{homayounfar2017sports,chen2019sports}, 
TS-WorldCup \citep{chu2022sports}, 
SoccerNet-Calibration \citep{cioppa2022scaling}, 
and CARWC \citep{claasen2024video} 
for soccer, or DeepSportRadar and 3DMPB \citep{van2022deepsportradar,huang2022pose2uv}, 
and the College Basketball Dataset \citep{sha2020end}, 
have been used for research. These datasets support the development of robust detection models across various sports. However, larger amount and diverse datasets require more annotation costs for enhancing model generalizability.

Recent advances in computer vision have highlighted the potential of synthetic datasets due to their efficiency in generation, diversity, and annotation quality. These datasets can be classified into two types: those synthesized from real-world images and those generated using simulators. In the field of computer vision, synthetic datasets derived from real images often involve techniques such as image augmentation and GAN (Generative Adversarial Network) \citep{zhao2020image} to enhance diversity. Image augmentation involves applying random transformations, such as rotation, flipping, or color adjustments, to increase the variability of training data without collecting new images. GANs improve the generator's ability to create realistic data by training it in an adversarial process against a discriminator, which tries to distinguish between real and generated data. Simulator-based datasets, such as those generated using Unreal Engine \citep{10171761}, enable highly controllable environments and provide detailed annotations, making them ideal for complex tasks like player tracking and field detection.

In soccer research, simulators (e.g., \cite{kurach2020google,liu2019emergent,kitano1998robocup}) have been widely utilized. However, existing soccer simulators are limited in their ability to generate bounding boxes for player detection. To address this limitation, Qin et al. developed the SoccerSynth simulator \citep{qin2025soccersynthdetectionsyntheticdatasetsoccer}.

However, the challenge remains in demonstrating their generalizability across other real-world tasks such as field detection. These datasets allow for controlled environments and the ability to simulate different scenarios that may be rare or hard to capture in real-world datasets. This makes synthetic datasets a promising alternative or supplement to real-world data, especially in pre-training models for tasks like soccer field detection.

In this study, we examine how using synthetic datasets as a supplement to real-world datasets during the training process can enhance the final performance of the model.
Field detection methods in sports analytics can be broadly categorized into keypoint-based approaches and keypoint-less approaches \citep{fujii2025machine}. Keypoint-based approaches focus on detecting specific field features, such as line intersections and center circles, to estimate homography and map the field to an overhead view. This category includes traditional methods based on RANSAC (RANdom SAmple Consensus) and DLT (Direct linear transformation) algorithms \citep{derpanis2010overview,hatze1988high}, as well as deep learning approaches that predict keypoints directly from images. Keypoint-less approaches, on the other hand, do not rely on specific field landmarks but instead utilize field texture, edge alignment, or heatmaps to estimate the field layout. Notable examples include methods like TVCalib \citep{theiner2023tvcalib}, which leverage entire field segmentations for accurate field detection. 

While existing methods have improved field detection accuracy and dataset diversity, they often rely on fixed settings and lack flexibility in randomization, limiting real-world generalization. Our study addresses these gaps by introducing a more flexible synthetic dataset generation framework. By pre-training models with synthetic data and fine-tuning them with real-world data, we aim to assess whether the synthetic dataset can enhance the overall detection accuracy and robustness of the models. Our hypothesis is that models pre-trained on synthetic data will perform better due to the increased diversity and control over the data.

 The contributions of this paper are as follows: 
    \begin{enumerate}
      \item We provide a novel approach to generate a synthetic soccer field dataset using the simulator \citep{qin2025soccersynthdetectionsyntheticdatasetsoccer} based on Unreal Engine 5, which serves as a flexible and efficient alternative to real-world data collection. Unlike existing soccer simulators, our approach allows dynamic adjustments to lighting, camera angles, and field elements.


      \item Our experimental results show that models pre-trained with the synthetic SoccerSynth-Field dataset outperform those trained solely on real-world data, contributing to better performance in soccer field detection tasks. This supports the potential of synthetic data as a cost-effective solution for improving model robustness in environments with limited real-world annotations.
    \end{enumerate}


\section*{Methods}
In this section, we describe the process used to generate the SoccerSynth-Field dataset, the model training procedure, 
and the evaluation metrics employed to assess the model's performance. The methods aim to explore the effectiveness of using a synthetic dataset for pretraining and then finetuning on real-world data.

\subsection*{Simulator}

\begin{figure}[H]
  \centering
  \includegraphics[width=0.8\textwidth, height=0.4\textheight]{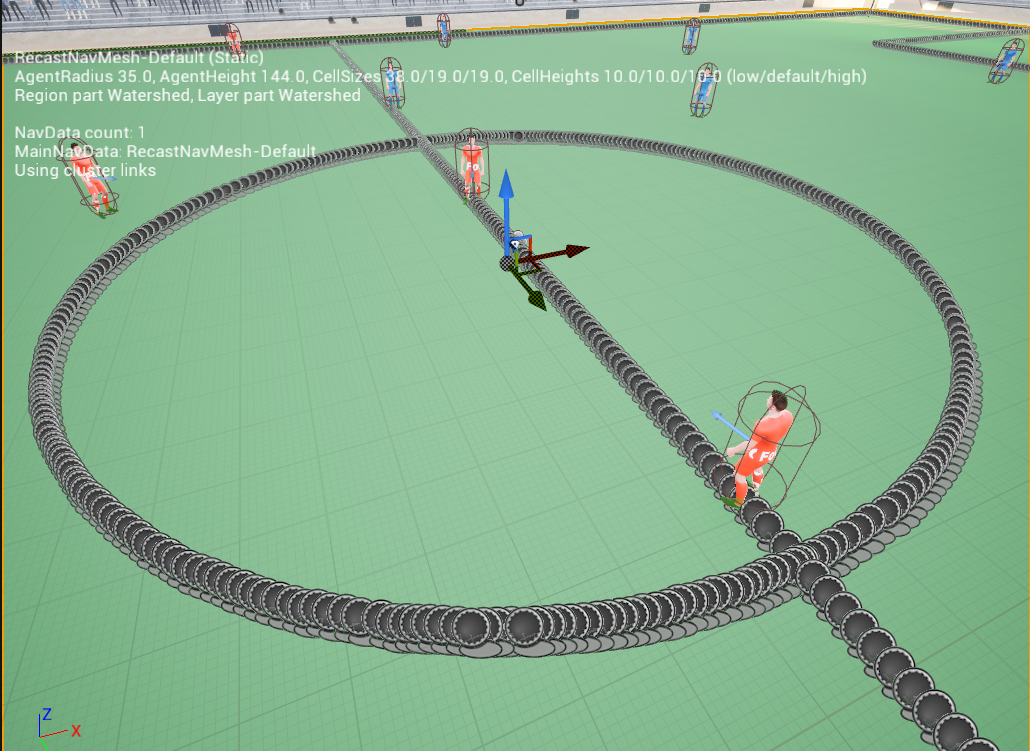} 
  \caption{Keypoints in our simulator: We placed numerous keypoints along the soccer field lines, naming them accordingly. During screenshots, we preserved the names and coordinates of these keypoints to reconstruct their trajectories. We then label spceific keypoints for training the model.} 
  \label{fig:kps_on_the_field} 
\end{figure}

The SoccerSynth-Detection Simulator \citep{qin2025soccersynthdetectionsyntheticdatasetsoccer}, developed using Unreal Engine, replicates soccer broadcasts to create a synthetic player detection dataset. We retained essential features of the SoccerSynth-Detection simulator, such as a combination of ten yellow and green grass colors and five grass textures, along with simulations of varying lighting conditions and diversified player appearances using grass textures and mappings to enrich dataset diversity. In this research, we introduced fake field lines in the simulator by randomly connecting two points on the soccer field's perimeter with a white line to add visual noise and simulate more authentic conditions.
Since the soccer pitch is a single texture map, we cannot separate the field lines for editing. To extract the field line information within the frame, we positioned numerous transparent keypoint instances named after the corresponding field lines, as illustrated in Figure \ref{fig:kps_on_the_field}. Using the engine's built-in functions, we are able to capture camera frames while preserving the names and coordinates of keypoints in all frames, used to reconstruct the trajectory of the field lines and generate the Soccersynth-Field dataset.
We then label spceific keypoints for training the model.

\begin{figure}[H]
    \centering
    \includegraphics[width=0.8\linewidth]{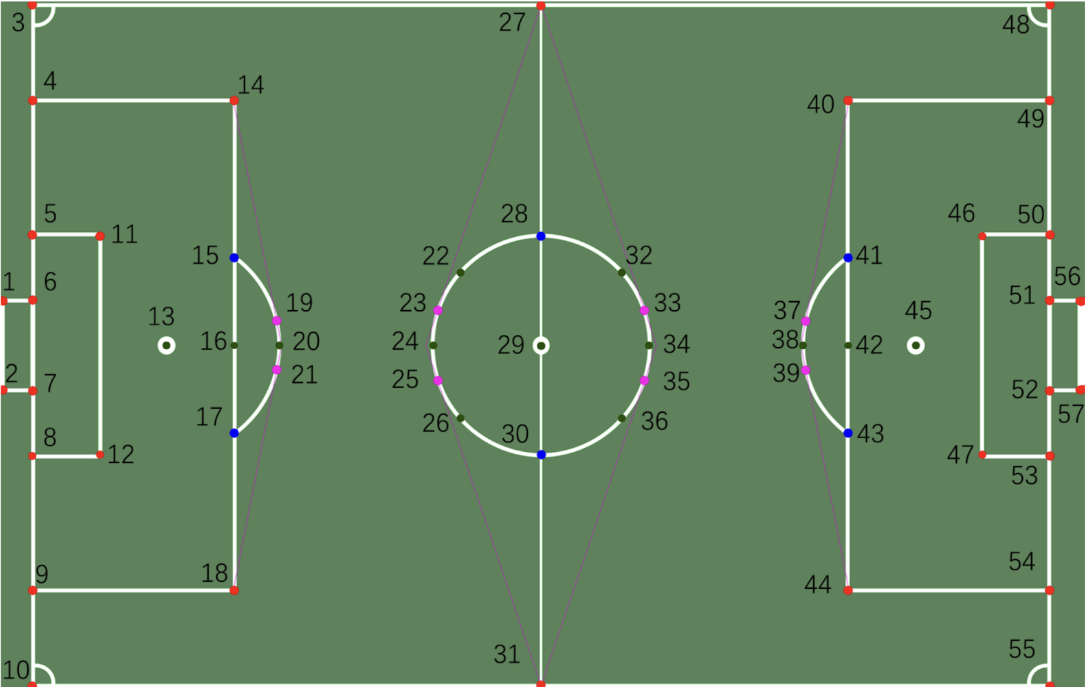}
    \caption{Tags of field. This figure illustrates the annotations of the soccer field such as side lines, goal areas, and center circles. The purpose is to provide a detailed reference for analyzing field elements in field detection.}
    \label{fig:field-tag}
\end{figure}

\subsubsection{Label Generation}\label{subsec1}
For each dataset, the images were resized to 1920 x 1080 pixels. Annotations for all datasets were provided in JSON format, containing the following information about the positions of the field lines and other relevant landmarks, as shown in Figure \ref{fig:field-tag}.

\begin{enumerate}
\item Big Rectangles:
Big rect. left bottom, Big rect. left main, Big rect. left top, 
Big rect. right bottom, Big rect. right main, Big rect. right top

\item Central Circle:
Circle central, Circle left, Circle right

\item Goals:
Goal left crossbar, Goal left post left, Goal left post right, 
Goal right crossbar, Goal right post left, Goal right post right

\item Middle Line:
Middle line

\item Side Lines:
Side line bottom, Side line left, Side line right, Side line top

\item Small Rectangles:
Small rect. left bottom, Small rect. left main, Small rect. left top, 
Small rect. right bottom, Small rect. right main, Small rect. right top
\end{enumerate}

These annotations were used as ground truth labels for training and evaluating the model's performance.
In total, the incremental complexity introduced across these datasets allowed us to systematically evaluate the impact of diversity in synthetic datasets on the effectiveness of pre-training models and fine-tuning them for real-world applications.

\subsection*{Generated Datasets}
We employed several distinct datasets for our experiments, each designed to incrementally introduce complexity and realism.
To improve the generalization of soccer field detection models, it is essential to simulate diverse real-world conditions using synthetic datasets. We designed 6 types of datasets, each incrementally introducing randomization elements, to observe their effects on model performance and robustness. These datasets aim to reflect real-world complexities such as player uniforms, audience presence, lighting conditions, and camera angles.

The SoccerSynth-Field dataset (final) serves as the complete dataset in this study. In this dataset, fake field lines were deliberately added to the images, introducing a significant challenge for the model to accurately detect and differentiate between real and artificial field lines. This setup tests the robustness of the model under conditions that closely resemble real-world scenarios with potential noise or ambiguities.

\begin{figure}[H]
    \centering
    \includegraphics[width=0.5\linewidth]{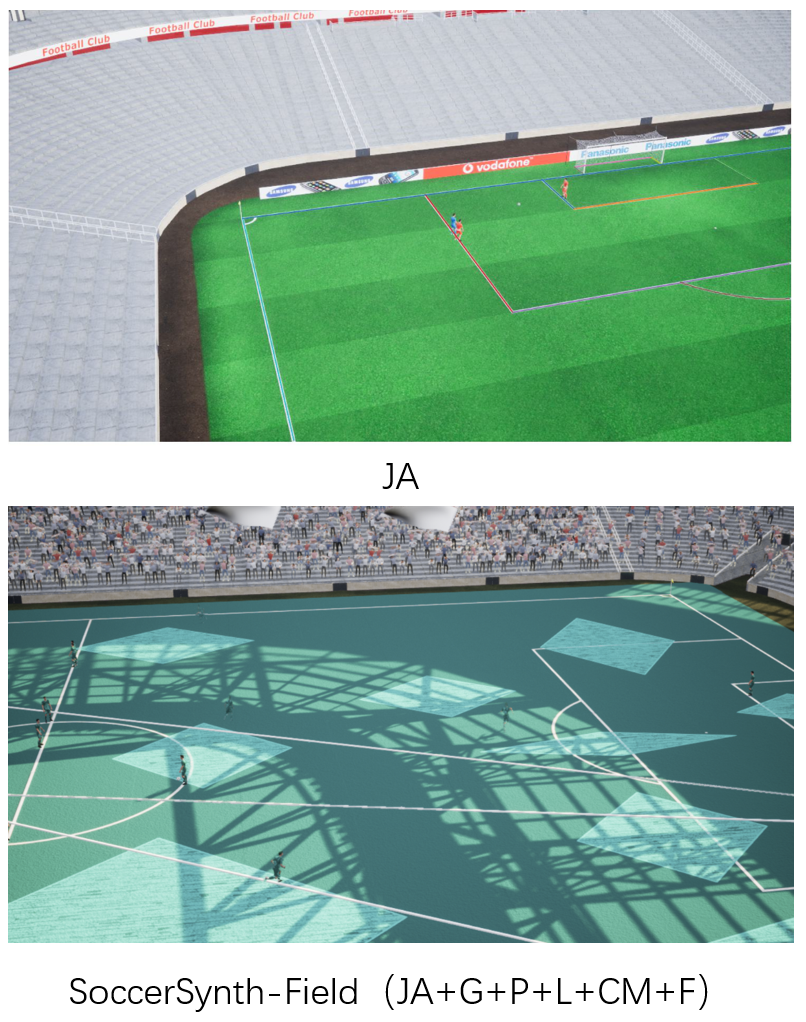}
    \caption{Comparison of synthetic datasets for soccer field detection: The upper image is JA dataset (audience and jersey randomizations only), which lacks complex environmental factors. The lower image is SoccerSynth-Field (JA+G+P+L+CM+F) dataset, added elements such as lighting, ground patterns, and audiences, enhanced diversity and realism to improve model robustness and generalization to real-world conditions.}
    \label{fig:eg1}
\end{figure}

To evaluate the model's performance under varying levels of complexity and realism, several other datasets were created for comparison (Figure \ref{fig:eg1} presents several examples):

\begin{enumerate}      
      \item JA dataset: This dataset represents the simplest setup. Audience randomization (A) was included, introducing variations in the presence and distribution of spectators. Additionally, jersey randomization (J) was applied, varying the colors of players’ uniforms to test the model's ability to adapt to these changes. All other parameters, such as camera angles, positions, and ground color, were fixed to maintain a controlled environment.

      \item JA+G dataset: Building upon the structured JA dataset, the JA+G dataset introduced greater variability. It includes jersey randomization (J), audience randomization (A), and introduces green randomization (G), which applies varying shades of green to the field to simulate natural variations in grass color. This tests the model's robustness to subtle but realistic changes in field appearance. All other parameters, such as camera positions and lighting conditions, remain fixed, ensuring that the dataset isolates these three factors for focused evaluation.

      \item JA+G+P dataset: This dataset extends the JA+G Dataset by incorporating ground pattern randomization (P), adding visual complexity to the field with designs commonly seen in professional soccer stadiums, challenging the model's ability to detect and analyze the target features.

      \item JA+G+P+L dataset: This dataset builds upon the JA+G+P dataset by adding lighting color adjustment (L), which simulates natural lighting variations. The lighting alternates between white and yellow hues, mimicking different times of the day and match conditions. Alongside jersey randomization (J), audience randomization (A), green randomization (G), and ground pattern randomization (P), this dataset introduces environmental diversity to further challenge the model. Other parameters, such as camera positions, remain fixed to isolate the effects of these factors.

      \item JA+G+P+L+CM dataset: This dataset extends the JA+G+P+L dataset by incorporating multiple cameras (M) and ground color randomization (C), providing varied perspectives and further increasing the dataset’s complexity. In addition to jersey randomization (J), audience randomization (A), green randomization(G), ground pattern randomization (P), and lighting color adjustment (L), this dataset achieves greater realism and diversity. The inclusion of multiple camera viewpoints enhances the model's ability to adapt to different angles and perspectives, making it more robust for real-world applications.
      
      \item SoccerSynth-Field dataset (final): In the SoccerSynth-Field dataset, fake field lines (F) were added to the images. This introduced additional challenges for the model, as it needed to accurately detect and differentiate between real and artificial lines, thereby testing its robustness against such synthetic modifications.

\end{enumerate}

The selection of these six datasets was based on balancing complexity and coverage. By progressively adding randomization factors, we ensure that each element's individual and combined effects are evaluated. Fewer than six types would miss key randomization combinations (e.g., lighting color or ground color), while additional variations may introduce unnecessary redundancy or overly specific conditions not commonly encountered in real broadcasts.

\subsection*{Training}

In the training phase, our goal was to assess the impact of synthetic data pretraining and real-world finetuning on model performance. We utilized the datasets described in the previous sections, the synthetic dataset contains 20,000 images to ensure sufficient diversity without causing excessive computational burden. In addition to these generated datasets, we incorporated real-world data from the SoccerNet-Calibration dataset, selecting 14,512 images and associated JSON files as our baseline training dataset. The SoccerNet-Calibration dataset was selected due to its high-quality annotations and widespread use in related studies. 

The segmentation model selected was DeepLabV3 with a ResNet-101 backbone, configured with auxiliary loss for intermediate representation learning. We used a batch size of 8, an initial learning rate of 0.01 with a polynomial decay, SGD optimizer (momentum 0.9, weight decay 1e-4), and a 30-epoch training schedule.

Next, we trained a model solely on 20,000 synthetic images to evaluate how well the model performs when trained exclusively on synthetic data. Finally, we finetuned the synthetic model with real-world data to bridge the domain gap and assess improvements from this hybrid approach.

For testing, we employed 186 images and JSON files from the WC14 dataset to evaluate the model's performance.

Our model training leveraged the codebase provided by TVcalib \citep{theiner2023tvcalib}. While we referenced TVcalib, our primary focus is on assessing the impact of synthetic datasets on field detection accuracy. Therefore, our main emphasis lies within the sn-segmentation component.

To this end, we conducted the following experiments:

\begin{enumerate}
\item To establish a reliable reference point, we first trained the sn-segmentation model using 14,512 real-world images from the SoccerNet-Calibration dataset. This serves as our baseline model to compare against the performance of models pre-trained on synthetic data.

\item Next, we trained a second model (Model 2) using only 20,000 images from the synthetic dataset.

\item Finally, we finetuned Model 2 using the 14,512 real-world images, producing our model (Model 3).

\end{enumerate}

Each of these models was subsequently evaluated on the WC14 dataset to compare their accuracy. Our main focus is on the comparison between Model 3 and the baseline, as we anticipate that training solely on synthetic data may not yield optimal results.







\section*{Results}
The primary goal of this experiment was to evaluate whether pretraining with synthetic datasets improves model performance in terms of robustness and accuracy in real-world conditions. The following sections provide quantitative results and qualitative comparisons to validate this hypothesis.

First, we present the results of our experiments, starting with a baseline model trained solely on real-world data using the sn-segmentation model. This baseline provides a benchmark against which we evaluate models pre-trained on synthetic datasets. Each synthetic dataset was used to pre-train a model, named according to the dataset it was trained on for clarity. 
Table \ref{tab:table1} indicates the accuracy achieved by each model before and after finetuning.

\begin{table}[H]
  
  \centering
  \resizebox{\textwidth}{!}{
    \begin{tabular}{ccc}
      \toprule
      Model  & Acc (\%) & Acc after finetune (\%) \\
      \midrule
      baseline (without synthetic data pretraining)         & - & 98.3  \\
      JA+G+P+L+CM+F (SoccerSynth-Field)       & 95.7      & \bf{98.5} \\
      JA+G+P+L+CM       & 95.5      & 98.4      \\
      JA+G+P+L       & 95.4      & 98.4      \\
      JA+G+P       & 94.2      & 98.4      \\
      JA+G       & 93.6      & 98.4     \\
      JA       & 92.6      & 97.5     \\
      \bottomrule
    \end{tabular}
  }
   \caption{Performance of the model using different datasets. Accuracy (Acc).\\
    \textit{Note}: The reported values indicate no standard deviation, because the three seeds produced almost consistent results across all tests. Therefore, a fixed value is presented as the average.}
    \label{tab:table1}
\end{table}

In the left column of Table \ref{tab:table1}, each model was initially pretrained on a specific synthetic dataset, with the pre-training accuracy presented. Among the synthetic models, JA+G+P+L+CM and SoccerSynth achieved the highest pretraining accuracy, with scores around 95.6\% and 95.7\%. These values, though lower than the baseline, suggest that incorporating more complex scenes and additional elements, such as varied lighting and camera perspectives, has a positive impact on the models' ability to generalize within the synthetic environment. Comparatively, simpler datasets, such as JA and JA+G, yielded lower pretraining accuracy, indicating a more limited capacity for realistic field detection.
We also compared the effectiveness of finetuning between JA, SoccerSynth Field, and the baseline model. The results confirmed the effectiveness of synthetic datasets in improving field detection in soccer broadcasts. 

Subsequently, we finetuned each pretrained model using real-world data, with results in the right column of Table \ref{tab:table1}. Finetuning consistently improved the accuracy across all models. Most synthetic pre-trained models achieved accuracy levels comparable to the baseline model's 98.3\%.  Notably, the SoccerSynth model outperformed the baseline slightly after finetuning, reaching 98.5\%.
In contrast, simpler datasets, such as JA (97.5\%), showed lower post-finetuning accuracy, suggesting that limited randomization may restrict generalization capabilities.

Next, Table \ref{tab:table2} shows that models pretrained with various amounts of synthetic data (SoccerSynth-Field+finetune) consistently achieve equal or superior performance compared to the same model trained only on real-world data. Notably, at the 9000 frames, the SoccerSynth+finetune model achieves 98.3\%,  which is 0.5\% higher than the baseline (97.8\%). As the dataset size decreases, the advantage of synthetic pretraining becomes more evident, maintaining robust performance even with smaller datasets (9000 and 6000 frames). However, at 6000 frames, the performance improvement was not as significant as at 9000 frames, likely due to insufficient finetuning samples at the smaller dataset size. These results highlight the effectiveness of synthetic data in improving generalization, especially in low-data scenarios.

\begin{table}[H]
\centering
\small

\begin{tabular}{ccc}
\toprule
\multicolumn{1}{c}{Trainig frames} & \multicolumn{2}{c}{Test Accuracy (\%)} \\ 
\cmidrule(lr){2-3}  
~~~(SoccerNet) & Without Pretrain & With Pretrain \\ 
\midrule
~~~~~~~14000       & 98.3 & \textbf{98.5}  \\ 
~~~~~~~~9000     & 97.8  & \textbf{98.3}\\ 
~~~~~~~~6000     & 97.8 &  \textbf{98.1}\\ 
\bottomrule
\end{tabular}
   \caption{Performance comparison of models trained on only real-world datasets (without pretrain) and models pre-trained on synthetic data (SoccerSynth) followed by fine-tuning on real-world datasets. The table highlights the accuracy achieved with different data scales (14000, 9000, and 6000 samples), demonstrating the effectiveness of synthetic data pretraining in improving model performance, particularly with smaller datasets.}
\label{tab:table2}

\end{table}

\begin{figure}[H]
    \centering
    \includegraphics[width=0.5\linewidth]{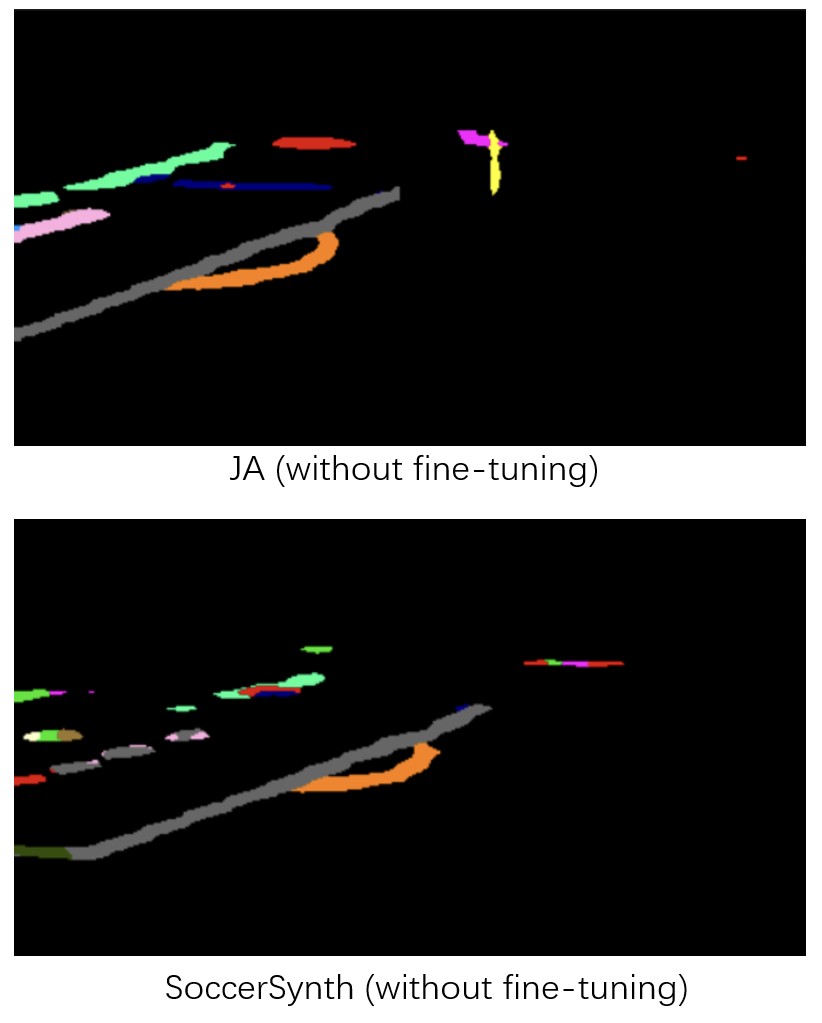}
    \caption{Comparison of field detection performance in JA (upper) and SoccerSynth-Field (lower) dataset without finetuning. Although the JA shows frequent misdetections, the SoccerSynth-Field demonstrates higher accuracy due to its exposure to diverse randomization elements during training. }
    \label{fig:JA}
\end{figure}

Furthermore, we show qualitative results.
To illustrate the impact of SoccerSynth dataset diversity, we provide the following visual comparisons between JA and SoccerSynth without finetuning in Figure \ref{fig:JA}.
The JA model struggles with key line detection in complex scenes, showing frequent misdetections and omissions.
In contrast, the SoccerSynth model, despite not being finetuned, demonstrates relatively higher initial accuracy due to its diverse randomization, although it still requires finetuning to achieve robust generalization in real-world scenarios.

\begin{figure}[H]
    \centering
    \includegraphics[width=0.5\linewidth]{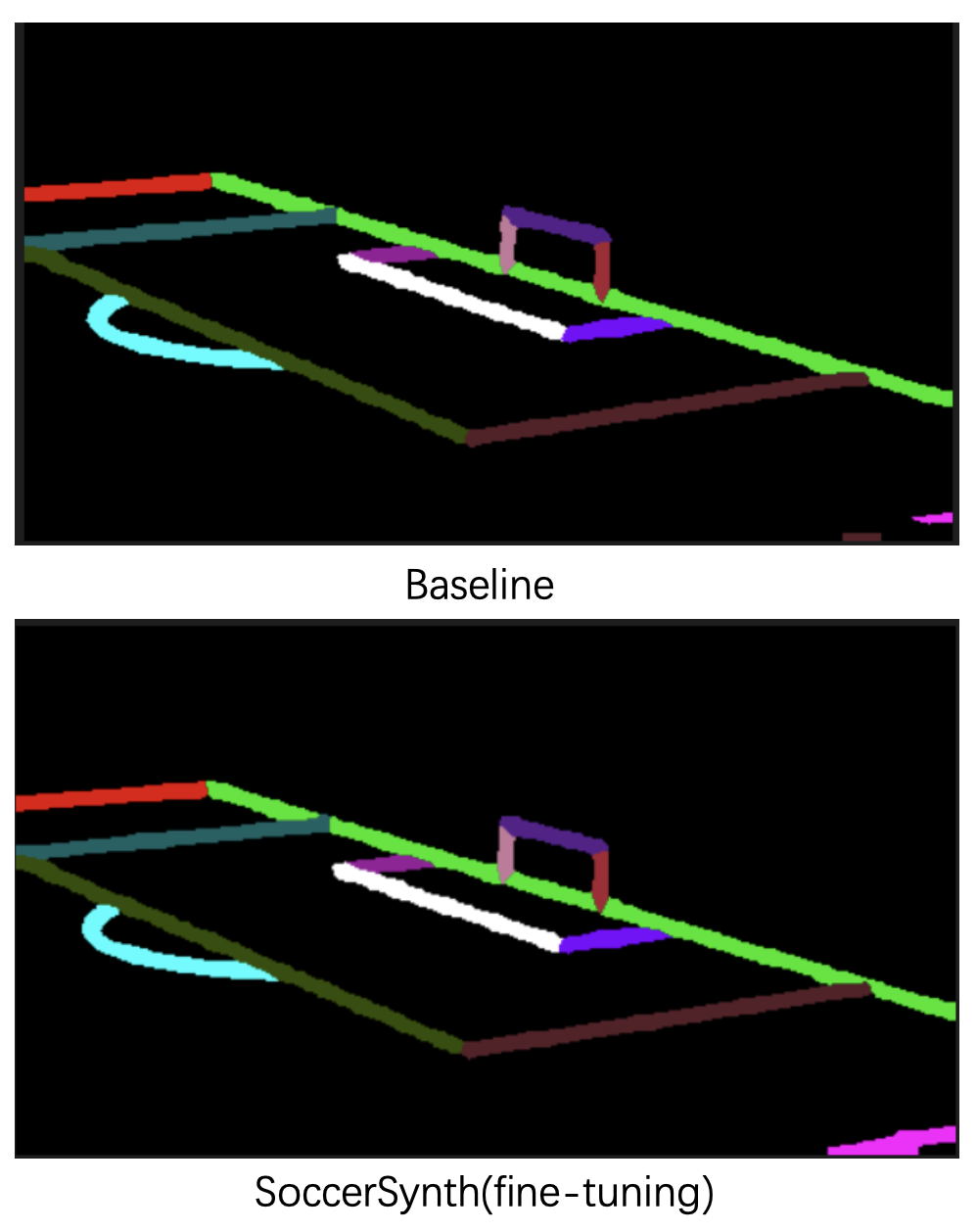}
    \caption{Comparison of baseline (real-world data only) and  SoccerSynth-Field(with pretraining). Both of them achieve comparable detection accuracy, even in scenes with noise and occlusions. Moreover, the SoccerSynth-Field model excels in handling finer details, as exemplified in the bottom-right corner of the figure, where it avoids the errors observed in the baseline model, indicating that synthetic data pretraining can effectively complement real-world data without causing noticeable degradation in results.}
    \label{fig:Baseline}
\end{figure}

Lastly, to show the impact of pretraining with our dataset, we show the comparison between the models with and without pretraining using our dataset in Figure \ref{fig:Baseline}.
The baseline model, trained solely on real-world data, serves as a benchmark for evaluating the performance improvements introduced by pre-training with synthetic data.  
The baseline model and the finetuned SoccerSynth model perform similarly in most scenarios, achieving accurate field detection.
However, the SoccerSynth model achieves this performance with the advantage of using pretraining on synthetic data, demonstrating that synthetic data can be an effective substitute when real-world data is scarce.
This improvement implies that certain synthetic datasets, especially those with features approximating real-world variability (e.g., lighting, camera angles), contribute meaningfully to model generalization for field recognition tasks.


These results indicate that using synthetic datasets for pretraining effectively improves model generalization. However, the effectiveness of specific randomization elements requires further analysis, as simpler datasets showed lower pretraining accuracy.

\section*{Discussion}

The experimental results demonstrate that synthetic datasets can effectively support the training of models for field detection tasks in soccer broadcast videos. And synthetic datasets with more diverse randomization elements lead to better model performance after finetuning. For instance, the SoccerSynth-Field dataset incorporates multiple cameras, lighting variations, and artificial noise, enabling the model to adapt to real-world complexities more effectively. In contrast, simpler datasets, such as JA, lacked such variability, resulting in limited generalization capability. This finding reinforces the importance of incorporating diverse randomization elements to simulate real-world scenarios.

Compared to traditional approaches \citep{homayounfar2016soccer}, our synthetic dataset provides a flexible and diverse training environment. By introducing randomized lighting, camera angles, and fake field lines, the SoccerSynth-Field dataset simulates complex noise patterns found in real broadcasts, improving the model's robustness in field detection tasks. This demonstrates that synthetic data can be a valuable resource for pre-training, particularly in scenarios where real-world data collection is limited or expensive.

Despite the promising results, there are limitations to our approach. The synthetic datasets, while diverse, may not fully capture all the nuanced textures and environmental dynamics of real-world broadcasts. However, a key strength of our method is its flexibility to design synthetic datasets that replicate specific scenarios, even for datasets without annotations. By tailoring the data generation process, our approach can simulate various game conditions and visual features, thereby compensating for the lack of labeled real-world data and improving the model’s generalization in diverse environments.

In summary, synthetic datasets hold promise as complementary tools for training in field detection, with further refinements likely to enhance their effectiveness and applicability in real-world scenarios.
 
\section*{Conclusion}

In this paper, we demonstrated that synthetic datasets can enhance the robustness of soccer field detection models through pre-training. Our approach incorporated various randomization elements, such as lighting, color, and synthetic noise, which improved generalization to real-world conditions. The SoccerSynth-Field dataset, in particular, highlighted the importance of realistic variability, achieving the highest accuracy after finetuning.

Our research emphasized the potential of synthetic data as a cost-effective solution for pretraining when real-world data is scarce. Future work will further explore additional environmental variations to improve realism and robustness.

\bibliography{bib2020}

@inproceedings{theiner2023tvcalib,
  title={Tvcalib: Camera calibration for sports field registration in soccer},
  author={Theiner, Jonas and Ewerth, Ralph},
  booktitle={Proceedings of the IEEE/CVF Winter Conference on Applications of Computer Vision},
  pages={1166--1175},
  year={2023}
}

@inproceedings{kitano1998robocup,
  title={The RoboCup synthetic agent challenge 97},
  author={Kitano, Hiroaki and Tambe, Milind and Stone, Peter and Veloso, Manuela and Coradeschi, Silvia and Osawa, Eiichi and Matsubara, Hitoshi and Noda, Itsuki and Asada, Minoru},
  booktitle={RoboCup-97: Robot Soccer World Cup I 1},
  pages={62--73},
  year={1998},
  organization={Springer},
  doi = "10.1007/3-540-64473-3_49"
}

@article{liu2019emergent,
  title={Emergent coordination through competition},
  author={Liu, Siqi and Lever, Guy and Merel, Josh and Tunyasuvunakool, Saran and Heess, Nicolas and Graepel, Thore},
  journal={arXiv preprint arXiv:1902.07151},
  year={2019},
  doi = "10.48550/arXiv.1902.07151"
}

@inproceedings{kurach2020google,
  title={Google research football: A novel reinforcement learning environment},
  author={Kurach, Karol and Raichuk, Anton and Sta{\'n}czyk, Piotr and Zaj{\k{a}}c, Micha{\l} and Bachem, Olivier and Espeholt, Lasse and Riquelme, Carlos and Vincent, Damien and Michalski, Marcin and Bousquet, Olivier and others},
  booktitle={Proceedings of the AAAI conference on artificial intelligence},
  volume={34},
  number={04},
  pages={4501--4510},
  year={2020},
  doi = "10.1609/aaai.v34i04.5878"
}

@INPROCEEDINGS{10171761,
  author={Damian, Alexandru and Filip, Claudiu and Nistor, Anamaria and Petrariu, Irina and Mariuc, Cătălin and Stratan, Valentin},
  booktitle={2023 17th International Conference on Engineering of Modern Electric Systems (EMES)}, 
  title={Experimental Results on Synthetic Data Generation in Unreal Engine 5 for Real-World Object Detection}, 
  year={2023},
  volume={},
  number={},
  pages={1-4},
  doi={10.1109/EMES58375.2023.10171761}}

@inproceedings{homayounfar2017sports,
  title={Sports field localization via deep structured models},
  author={Homayounfar, Namdar and Fidler, Sanja and Urtasun, Raquel},
  booktitle={Proceedings of the IEEE Conference on Computer Vision and Pattern Recognition},
  pages={5212--5220},
  year={2017}
}

@inproceedings{chen2019sports,
  title={Sports camera calibration via synthetic data},
  author={Chen, Jianhui and Little, James J},
  booktitle={Proceedings of the IEEE/CVF conference on computer vision and pattern recognition workshops},
  pages={0--0},
  year={2019}
}

@inproceedings{chu2022sports,
  title={Sports field registration via keypoints-aware label condition},
  author={Chu, Yen-Jui and Su, Jheng-Wei and Hsiao, Kai-Wen and Lien, Chi-Yu and Fan, Shu-Ho and Hu, Min-Chun and Lee, Ruen-Rone and Yao, Chih-Yuan and Chu, Hung-Kuo},
  booktitle={Proceedings of the IEEE/CVF Conference on Computer Vision and Pattern Recognition},
  pages={3523--3530},
  year={2022}
}

@article{cioppa2022scaling,
  title={Scaling up SoccerNet with multi-view spatial localization and re-identification},
  author={Cioppa, Anthony and Deliege, Adrien and Giancola, Silvio and Ghanem, Bernard and Van Droogenbroeck, Marc},
  journal={Scientific data},
  volume={9},
  number={1},
  pages={355},
  year={2022},
  publisher={Nature Publishing Group UK London}
}

@article{claasen2024video,
  title={Video-based sequential Bayesian homography estimation for soccer field registration},
  author={Claasen, Paul Johannes and De Villiers, Johan Pieter},
  journal={Expert Systems with Applications},
  volume={252},
  pages={124156},
  year={2024},
  publisher={Elsevier}
}

@inproceedings{van2022deepsportradar,
  title={Deepsportradar-v1: Computer vision dataset for sports understanding with high quality annotations},
  author={Van Zandycke, Gabriel and Somers, Vladimir and Istasse, Maxime and Don, Carlo Del and Zambrano, Davide},
  booktitle={Proceedings of the 5th International ACM Workshop on Multimedia Content Analysis in Sports},
  pages={1--8},
  year={2022}
}

@article{huang2022pose2uv,
  title={Pose2uv: Single-shot multiperson mesh recovery with deep uv prior},
  author={Huang, Buzhen and Zhang, Tianshu and Wang, Yangang},
  journal={IEEE Transactions on Image Processing},
  volume={31},
  pages={4679--4692},
  year={2022},
  publisher={IEEE}
}

@inproceedings{sha2020end,
  title={End-to-end camera calibration for broadcast videos},
  author={Sha, Long and Hobbs, Jennifer and Felsen, Panna and Wei, Xinyu and Lucey, Patrick and Ganguly, Sujoy},
  booktitle={Proceedings of the IEEE/CVF conference on computer vision and pattern recognition},
  pages={13627--13636},
  year={2020}
}

@book{fujii2025machine,
  title={Machine Learning in Sports:
Open Approach for Next Play
Analytics},
  author={Fujii, Keisuke},
  year={2025},
  publisher={Springer}
}

@article{derpanis2010overview,
  title={Overview of the RANSAC Algorithm},
  author={Derpanis, Konstantinos G},
  journal={Image Rochester NY},
  volume={4},
  number={1},
  pages={2--3},
  year={2010}
}

@article{hatze1988high,
  title={High-precision three-dimensional photogrammetric calibration and object space reconstruction using a modified DLT-approach},
  author={Hatze, Herbert},
  journal={Journal of biomechanics},
  volume={21},
  number={7},
  pages={533--538},
  year={1988},
  publisher={Elsevier}
}

@article{zhao2020image,
  title={Image augmentations for gan training},
  author={Zhao, Zhengli and Zhang, Zizhao and Chen, Ting and Singh, Sameer and Zhang, Han},
  journal={arXiv preprint arXiv:2006.02595},
  year={2020}}

@article{homayounfar2016soccer,
  title={Soccer field localization from a single image},
  author={Homayounfar, Namdar and Fidler, Sanja and Urtasun, Raquel},
  journal={arXiv preprint arXiv:1604.02715},
  year={2016}
}

@misc{qin2025soccersynthdetectionsyntheticdatasetsoccer,
      title={SoccerSynth-Detection: A Synthetic Dataset for Soccer Player Detection}, 
      author={Qin, Haobin  and Yeung, Calvin  and Umemoto, Rikuhei  and  Fujii, Keisuke},
      year={2025},
      eprint={2501.09281},
      archivePrefix={arXiv},
      primaryClass={cs.CV},
      url={https://arxiv.org/abs/2501.09281}, 
}

\end{document}